\def\year{2017}\relax
\documentclass[letterpaper]{article}
\usepackage{aaai17}
\usepackage{times}
\usepackage{helvet}
\usepackage{graphicx}
\usepackage{courier}
\usepackage{amsmath}
\usepackage{amsfonts}
\usepackage{algorithm}
\usepackage{algorithmic}
\usepackage{multirow}
\usepackage{url}
\begin{document}
\title{Efficient Hyperparameter Optimization of Deep Learning Algorithms Using Deterministic RBF Surrogates}
\author{
Ilija Ilievski \\ Graduate School for Integrative Sciences and Engineering \\ National University of Singapore \\ \texttt{ilija.ilievski@u.nus.edu}
\And
Taimoor Akhtar \\ Industrial and Systems Engineering \\ National University of Singapore \\ \texttt{erita@nus.edu.sg}
\AND
Jiashi Feng \\ Electrical and Computer Engineering \\ National University of Singapore \\ \texttt{elefjia@nus.edu.sg}
\And
Christine Annette Shoemaker \\ Industrial and Systems Engineering \\ National University of Singapore \\ \texttt{isesca@nus.edu.sg}
}
\maketitle

\begin{abstract}
Automatically searching for optimal hyperparameter configurations  is of crucial importance for applying deep learning algorithms in practice.
Recently, Bayesian optimization has been  proposed for optimizing hyperparameters of various machine learning algorithms.
Those methods adopt probabilistic surrogate models like Gaussian processes to approximate and minimize the validation error  function of hyperparameter values.
However, probabilistic surrogates require accurate  estimates of sufficient statistics (\emph{e.g.}, covariance) of the error distribution and thus need many function evaluations with a sizeable number of hyperparameters.
This makes them inefficient  for optimizing hyperparameters  of deep learning algorithms, which are highly  expensive to evaluate.
In this work, we propose a new deterministic and efficient hyperparameter optimization method that employs  radial basis functions as error   surrogates. 
The proposed mixed integer algorithm, called HORD,  searches the surrogate for the most promising hyperparameter values through dynamic coordinate search and requires many fewer function evaluations.
HORD~does well in low dimensions but it is exceptionally better in higher dimensions.
Extensive evaluations on MNIST and CIFAR-10 for four deep neural networks demonstrate HORD significantly outperforms the well-established Bayesian optimization methods such as GP, SMAC and TPE.
For instance, on average, HORD is more than  $6$  times faster than GP-EI in obtaining the  best configuration of $19$ hyperparameters.
\end{abstract}

\section{Introduction}\label{sec:intro}

Deep learning algorithms have been extensively used for various artificial intelligence related problems in recent years.
However, their adoption is severely hampered by the many hyperparameter choices one must make, \emph{e.g.}, architectures of the deep neural networks,  forms of the activation functions and learning rates.
Determining appropriate values of these hyperparameters is  of crucial importance  which however is a frustratingly difficult task.
As a result, automatic optimization of  hyperparameters for deep learning algorithms  has been attracting much research effort~\cite{bergstra2011algorithms,snoek2012practical,bergstra2013making,swersky2014freeze,snoek2015scalable}.

Hyperparameter optimization is a global optimization of a black-box error function $f$ whose evaluation is expensive~\textemdash~the function $f$ maps a hyperparameter choice $\mathbf{x}$ of $D$ configurable hyperparameters to  {validation} error  of a deep learning algorithm with learned parameters $\theta$.
Optimizing $f$ as follows  gives a way to automatically search for  optimal hyperparameters:
\begin{equation}
\label{eqn:hyp_opt_def}
\begin{aligned}
\min_{\mathbf{x}\in \mathbb{R}^D }\quad &f(\mathbf{x},\theta;\mathcal{Z}_{\mathrm{val}}), \\
\text{s.t. } \quad &\theta = \arg\min_{\theta} f(\mathbf{x}, \theta;\mathcal{Z}_{\mathrm{train}})
\end{aligned}
\end{equation}
Here $\mathcal{Z}_\mathrm{train}$ and $\mathcal{Z}_\mathrm{val}$ denote the training and validation datasets respectively,   $\theta$ is learned through minimizing the training error and $\mathbf{x}$ is in a bounded set. 
In the following, we omit (in the notation only) the  dependency of   $f$ on $\theta$ and $\mathcal{Z}$, and denote it as $f(\mathbf{x})$ for short.

Solving the problem in Eq.~\eqref{eqn:hyp_opt_def} is very challenging due to  high complexity of the function $f$.
A popular  solution   is to employ Bayesian optimization algorithms~\textemdash~which use  a cheap \emph{probabilistic} surrogate model to approximate the expensive error function~\cite{mockus1978application}, such as Gaussian process (GP)~\cite{snoek2012practical} and Tree-structured Parzen Estimator (TPE)~\cite{bergstra2011algorithms}.

In this work, we propose a new \emph{deterministic}-surrogate based hyperparameter optimization method  that we show requires considerably fewer function evaluations for the optimization.
The proposed algorithm uses radial basis function (RBF) as surrogate to
approximate the  error function  of hyperparameters, and avoids the expensive computation of covariance statistics needed in GP methods.
The proposed algorithm searches the surrogate via dynamic hyperparameter coordinate search and is able to find a near optimal hyperparameter configuration with few expensive function evaluations, for either integer or continuous valued hyperparameters. 

We name our algorithm as Hyperparameter Optimization via RBF and Dynamic coordinate search, or HORD for short.
We compare our method against the well-established   GP-based algorithms (\emph{i.e.}, GP-EI and GP-PES) and against the tree-based algorithms (\emph{i.e.}, TPE and SMAC)  for optimizing the hyperparameters of two types of commonly used neural networks applied on the MNIST and CIFAR-10 benchmark datasets.
	
Our contributions can be summarized as follows:
\begin{itemize}
	\item We provide a mixed integer deterministic-surrogate optimization algorithm for optimizing hyperparameters.
      The algorithm is capable of optimizing both continuous and integer hyperparameters of deep neural networks and performs equally well in low-dimensional hyperparameter spaces and exceptionally well in higher dimensions.
   \item Extensive evaluations demonstrate the superiority of the proposed algorithm over the state-of-the-art, both in finding a near-optimal configuration with fewer function evaluations and in achieving a lower final validation error. As discussed later, HORD obtains speedups of $3.7$ to $6$ fold over average computation time of other algorithms with $19$-dimensional problem and is faster on average for all dimensions and all problems  tested.
\end{itemize}

\section{Related Work}\label{sec:related_work}

Surrogate-based optimization~\cite{mockus1978application} is a strategy for the global optimization of expensive black-box functions over a constrained domain.
The goal is to obtain a near optimal solution with as few as possible function evaluations.
The main idea is to utilize a surrogate model of the expensive function that can be inexpensively evaluated to determine the next most promising point for evaluation.
Surrogate-based optimization methods differ in two aspects: the type of model used as a surrogate and the way the model is used to determine the next most promising point for expensive evaluation.

Bayesian optimization is a type of surrogate-based optimization, where the surrogate is a probabilistic model to compute the posterior estimate of distribution of the expensive error function values.
The next most promising point is determined by optimizing an acquisition function of choice given the distribution estimation.
Various  Bayesian optimizations methods mainly differ in how to estimate the error distribution and definition on the acquisition function.

GP-EI~\cite{snoek2012practical} and GP-PES~\cite{hernandez2014predictive}   methods use Gaussian processes to estimate  distribution of the validation error given a hyperparameter configuration using the history of observations.
The methods  use the expected improvement (EI) and  the predictive entropy search (PES) respectively as an acquisition function.
Although GP is simple and flexible, GP-based optimization involves inverting an expensive covariance matrix and thus scales cubically with the number of observations.
To address the scalability issue of GP based methods, TPE is proposed by~\cite{bergstra2011algorithms}.
TPE is a non-standard Bayesian-based optimization algorithm, which uses tree-structured Parzen estimators to model the error distribution in a non-parametric way.
SMAC~\cite{hutter2011sequential} is another tree-based algorithm that uses random forests to estimate the error density.

In~\shortcite{eggensperger2013towards}~\citeauthor{eggensperger2013towards}~empirically showed that Spearmint’s GP-based approaches reach state-of-the-art performance in optimizing few hyperparameters, while TPE's and SMAC's tree-structured approaches achieve best results in high-dimensional spaces.
Thus, we compare our algorithm against these four methods.

The authors of~\cite{eggensperger2015efficient} propose an interesting way of using surrogate based models for hyperparameter optimization. They propose to build a surrogate of the validation error given a hyperparameter configuration to \textit{evaluate and compare} different hyperparameter optimization algorithms. However, they do not unify the surrogate models with an optimization algorithm. 

In~\cite{snoek2015scalable} the authors address the scalability issue of GB-based models and explore the use of neural networks as an alternative surrogate model. 
Their neural network based hyperparameter optimization algorithm uses only a fraction of the computational resources of GP-based algorithms, especially when the number of function evaluation grows to $2{,}000$. 
However, as opposed to our proposed method, their approach does not outperform the GP-based algorithms nor the tree-based algorithms such as TPE and SMAC.

\section{Description of The Proposed Method HORD}

Here we introduce a novel hyperparameter optimization called Hyperparameter Optimization using RBF and Dynamic coordinate search, or HORD~for short.

\subsection{The surrogate model}

We use the radial basis function (RBF) interpolation model as the surrogate model~\cite{powell1990theory} in searching for optimal hyperparameters.
Given $D$ number of hyperparameters, and $n$ hyperparameter configurations, $\mathbf{x}_{1:n}$, with $\mathbf{x}_{i} \in \mathbb{R}^D$, and their corresponding validation errors $f_{i:n}$, where $f_{i} = f(\mathbf{x}_{i})$, we define the RBF interpolation model as:
{
	\begin{equation}
	\label{eqn:sur_def}
	S_{n}(\mathbf{x})=\sum_{i=1}^{n} \lambda_{i}\phi(\|\mathbf{x} - \mathbf{x_{i}}\|)+p(\mathbf{x})
	\end{equation}}

	Here $\phi(r) = r^3$ denotes the cubic spline RBF, $\| \cdot \|$ is the Euclidean norm, and $p(\mathbf{x})=\mathbf{b}^\top\mathbf{x} + a$ is the polynomial tail, with $\mathbf{b}=[b_{1},\ldots,b_{d}]^\top \in \mathbb{R}^D $, and $a \in \mathbb{R}$.
	The parameters $\lambda_{i}$, $i = 1,\ldots,n$, $b_{k}$, $k=1,\ldots,d$,  and $a$ are the interpolation model parameters, determined by solving the linear system of equations~\cite{gutmann2001radial}:

	{
		\begin{equation}
		\begin{bmatrix}
		\mathbf{\Phi} & \mathbf{P} \\
		\mathbf{P^\top} & \mathbf{0} \\
		\end{bmatrix}
		\begin{bmatrix}
		\mathbf{\lambda} \\
		\mathbf{c}  \\
		\end{bmatrix}
		=
		\begin{bmatrix}
		\mathbf{F} \\
		\mathbf{0}  \\
		\end{bmatrix}
		\end{equation}
	}
	
	Here elements in the matrix $\mathbf{\Phi} \in \mathbb{R}^{n\times n}$ are defined as $\mathit{\Phi_{i,j}}=\phi(\|\mathbf{x_{i}} - \mathbf{x_{j}}\|), i,j=1,\ldots,n$, $\mathbf{0}$ is a zero matrix of compatible dimension, and
	
	{
		\begin{equation}
		\mathbf{P} =  \begin{bmatrix} \mathbf{x_{1}}^\top  & 1 \\
		&\hspace{-8mm} \vdots \\
		\mathbf{x_{n}}^\top & 1 \\
		\end{bmatrix}
		\mathbf{\lambda} =
		\begin{bmatrix}
		\lambda_{1} \\
		\vdots \\
		\lambda_{n} \\
		\end{bmatrix}
		\mathbf{c} =
		\begin{bmatrix}
		b_{1}\\
		\vdots  \\
		b_{k} \\
		a \\
		\end{bmatrix}
		\mathbf{F} =
		\begin{bmatrix}
		f(x_{1})\\
		\vdots \\
		f(x_{n}) \\
		\end{bmatrix}
		\end{equation}
	}

\subsection{Searching the hyperparameter space}
The training and evaluation of a deep neural network (DNN) can be regarded as a function $f$ that maps the hyperparameter configuration used to train the network to the validation error obtained at the end.
Optimizing $f(\mathbf{x})$ with respect to a hyperparameter configuration $\mathbf{x}$ as in Eq.~\eqref{eqn:hyp_opt_def} is a global optimization problem since the function $f(\mathbf{x})$ is typically highly multimodal.
We employ the RBF interpolation to approximate the expensive function $f(\mathbf{x})$.
We build  the new \emph{mixed integer} global optimization HORD algorithm with some features from the continuous global Dynamic coordinate search (DYCORS-LMSRBF)~\cite{regis2013combining} to search the surrogate model for promising hyperparameter configurations.

As in the LMSRBF \cite{Regis2007c},  the DYCORS-LMSRBF algorithm starts by fitting an initial surrogate model $S_{n_0}$ using $A_{n_0}=\{(\mathbf{x}_i,f_i))\}_{i=1}^{n_0}$.
We set $n_{0}=2(D+1)$, and use the Latin hypercube sampling method to sample $\mathbf{x}_{1:n_0}$ hyperparameter configurations and obtain their respective validation errors $f_{1:n_0}$.

Next, while $n < N_{\max}$ a user-defined maximal evaluation number, the candidate point generation algorithm (Alg.~~\ref{alg:hord}) populates the candidate point set $\Omega_{n}={t_{n,1:m}}$, with $m=100D$ candidate points, and selects for evaluation the most promising point as $\mathbf{x}_{n+1}$.
After $N_{\max}$ iterations, the algorithm returns the hyperparameter configuration $\mathbf{x}_{\text{best}}$ that yielded the lowest validation error $f(\mathbf{x}_{\text{best}})$.
A formal algorithm description is given in Alg.~\ref{alg:hord}.

\begin{algorithm}
		\caption{\textbf{H}yperparameter \textbf{O}ptimization using \textbf{R}BF-based surrogate and \textbf{D}YCORS (HORD)}
	\begin{algorithmic}[1]
		\INPUT		$n_0=2(D+1)$, $m=100D$ and $N_{\max}$.
		\OUTPUT optimal hyperparameters $\mathbf{x}_{\text{best}}$. 
		\STATE Use Latin hypercube sampling to sample $n_0$ points and set $\mathcal{I} = \{\mathbf{x}_{i}\}_{i=1}^{n_0}$.
		\STATE Evaluating $f(x)$ for points in $\mathcal{I}$ gives $\mathcal{A}_{n_0} = \{(\mathbf{x}_{i}, f(\mathbf{x}_{i}))\}_{i=1}^{n_0}$.
		\WHILE{$n<N_{\max}$}
		\STATE Use $\mathcal{A}_{n}$ to fit or update the surrogate model $S_n(\mathbf{x})$ (Eq.~\ref{eqn:sur_def}).
		\STATE Set $\mathbf{x}_{\text{best}}=\arg \min\{f(\mathbf{x}_i): i=1,\ldots,n \}$.
      \STATE Compute $\varphi_n$ (Eq.~\ref{eq:varphi}), i.e, the probability of perturbing a coordinate.
		\STATE Populate $\Omega_{n}$ with $m$ candidate points, $\mathbf{t}_{n,1:m}$, where for each candidate $\mathbf{y}_{j} \in \mathbf{t}_{n,1:m}$, (a) Set $\mathbf{y}_{j} = \mathbf{x}_{\text{best}}$, (b) Select the coordinates of $\mathbf{y}_{j}$ to be perturbed with probability $\varphi_n$ and (c) Add $\delta_i$ sampled from $\mathcal{N}(0,\sigma_{n}^2)$ to the coordinates of $\mathbf{y}_{j}$ selected  in (b) and round to nearest integer if required.
      \STATE Calculate $V_{n}^{ev}(\mathbf{t}_{n,1:m})$ (Eq.~\ref{eqn:vev}), $V_{n}^{dm}(\mathbf{t}_{n,1:m})$ (Eq.~\ref{eqn:vdm}), and the final weighted score $W_n(\mathbf{t}_{n,1:m})$ (Eq.~\ref{eqn:weight}).
      \STATE Set $\mathbf{x}^{*} = \arg \min\{W_n(\mathbf{t}_{n,1:m})\}$.
		\STATE Evaluate $f(\mathbf{x}^{*})$.
		\STATE Adjust the variance $\sigma_{n}^2$ (see text).
		\STATE Update $\mathcal{A}_{n+1} = \{\mathcal{A}_{n} \cup (\mathbf{x}^{*},f(\mathbf{x}^{*}))\}$.
		\ENDWHILE
      \STATE Return $\mathbf{x}_{\text{best}}$.
	\end{algorithmic}
	\label{alg:hord}
\end{algorithm}

\subsection{Candidate hyperparameter generation and selection}\label{sec:cand_algo}

We describe how to generate and select  candidate hyperparameters in the following section.

The set of candidate hyperparameters $\Omega_{n}$, at iteration $n$ (Step 7 in Alg.~\ref{alg:hord}) is populated by points generated by adding  noise $\delta_i$ to some or all of the coordinates\footnote{A coordinate in the hyperparameter space, is one specific hyperparameter (\emph{e.g.}, learning rate).}  of the current $\mathbf{x}_{\text{best}}$.
In HORD, the expected number of  coordinates  perturbed  is monotonically decreasing since perturbing \textit{all} coordinates of the current $\mathbf{x}_{\text{best}}$ will result in a point much further from $\mathbf{x}_{\text{best}}$, a significant problem when the dimension $D$ becomes large, \emph{e.g.}, $D>10$.
Note that, during one iteration, each candidate point is  generated by perturbing a potentially different subset of coordinates.

 The probability of perturbing a coordinate (Step 6 in Alg.~\ref{alg:hord}) in HORD declines with the number of iterations and is given by:
 {
            \begin{equation}\label{eq:varphi}
            \varphi_n = \varphi_0\left[1 - { \ln(n-n_0 + 1) \over \ln(N_{\max} - n_0) }\right], \quad n_0<= n < N_{\max}
            \end{equation}
 }
 where $\varphi_0$ is set to $\min(20/D, 1)$ so that the average number of coordinates perturbed is always less than $20$. HORD combines features from DYCORS \cite{regis2013combining} and SO-MI \cite{Mueller2013} to generate candidates. DYCORS is a continuous optimization RBF Surrogate method that uses Eq.~\ref{eq:varphi} for controlling perturbations.  SO-MI is the first mixed integer implementation of a RBF surrogate method.

The perturbation $\delta_i$ is sampled from $\mathcal{N}(0,\sigma_n^2)$.
Initially, the variance $\sigma_{n_0}^2$ is set to $0.2$, and after each $\mathcal{T}_{\text{fail}}= \max(5,D)$ consecutive iterations with no improvement over the current $x_{\text{best}}$, is set to $\sigma_{n+1}^2 = \min(\sigma_{n}^2/2, 0.005)$.
The variance is doubled (capped at the initial value of $0.2$), after $3$ consecutive iterations with improvement.
The adjustment of the variance (Step 11 in Alg.~\ref{alg:hord}) is in place to facilitate the convergence of the optimization algorithm.
Once the set of candidate points $\Omega$ has been populated with $m=100D$ points, we estimate their corresponding validation errors by computing the surrogate values $S(\mathbf{t}_{1:m})$ (for clarity, we leave out $n$ from notation). Here $\mathbf{t}$ denotes a candidate point.

We compute the distances from the previously evaluated points $\mathbf{x}_{1:n}$ for each $\mathbf{t} \in \Omega$ with $\Delta(\mathbf{t}) = \min\| \mathbf{t} - \mathbf{x}_{1:n} \|$.
Here, $\| \cdot \|$ is the Euclidean norm.
We also compute $\Delta^{\max} = \max\{\Delta( \mathbf{t}_{1:m})\}$ and $\Delta^{\min} = \min\{\Delta( \mathbf{t}_{1:m})\}$.

Finally, for each $\mathbf{t} \in \Omega$ we compute the score for the two criteria: $V^{ev}$ for the surrogate estimated value:

{
   \begin{equation}
      \label{eqn:vev}
      V^{ev}(\mathbf{t}) =
      \begin{cases}
         \frac{S(t) - s^{\min}}{s^{\max}-s^{\min}}, &\text{if } s^{\max} \neq s^{\min};\\
         1, &\text{otherwise}.
      \end{cases}
   \end{equation}
}
And the distance metric $V^{dm}$:
{
   \begin{equation}
      \label{eqn:vdm}
V^{dm}(\mathbf{t}) =
\begin{cases}
\frac{\Delta^{\max} - \Delta(t)}{\Delta^{\max}-\Delta^{\min}}, &\text{if } \Delta^{\max} \neq \Delta^{min};\\
1, &\text{otherwise}.
\end{cases}
\end{equation} }
We use $V^{ev}$ and $V^{dm}$ to compute the final weighted score:
{
\begin{equation}
\label{eqn:weight}
W(\mathbf{t}) = wV^{ev}(\mathbf{t}) + (1-w)V^{dm}(\mathbf{t})
\end{equation}
}
The final weighted score (Eq.~\ref{eqn:weight}) is the acquisition function used in HORD to select a new evaluation point. Here, $w$ is a cyclic weight for balancing between global and local search. We cycle the following weights: $0.3, 0.5, 0.8, 0.95$ in sequential manner.
We select the point with the lowest weighted score $W$ to be the next evaluated point $\mathbf{x}_{n+1}$.

\citeauthor{Mueller2014b} \shortcite{Mueller2014b} compare the above candidate approach with instead searching with a genetic algorithm. \citeauthor{Krityakierne2016} \shortcite{Krityakierne2016} use multi-objective methods with RBF surrogate to select the next point for evaluation for single objective optimization of $f(x)$.
  
\section{Experimental Setup}\label{sec:experiments}
\paragraph{DNN problems} 

We test HORD on four DNN hyperparameter optimization problems with 6, 8, 15 and 19 hyperparameters. Our first DNN problem consists of $4$ continuous and $2$ integer hyperparameters of a Multi-layer Perceptron (MLP) network applied to classifying grayscale images of handwritten digits from the popular benchmark dataset MNIST. This problem is also referred as \textbf{6-MLP} in subsequent discussions.

The MLP network consists of two hidden layers with \emph{ReLU} activation between them and \emph{SoftMax} at the end. As learning algorithm we use Stochastic Gradient Descent (SGD) to compute $\theta$ in Eq.~\ref{eqn:hyp_opt_def}. The hyperparameters (denoted as $x$ in Eq.~\ref{eqn:hyp_opt_def}) we optimize with HORD and other algorithms include hyperparameters of the learning algorithm, the layer weight initialization hyperparameters, and network structure hyperparameters. Full details of the used datasets, the hyperparameters being optimized and their respective value ranges, as well as the values used as initial starting point (always required by SMAC) are provided in the supplementary materials.

The second DNN problem (referred as \textbf{ 8-CNN} in subsequent discussions), has $4$ continuous and $4$ integer hyperparameters of a more complex Convolutional Neural Network (CNN).
The CNN consists of two convolutional blocks, each containing one convolutional layer with batch normalization, followed by \emph{ReLU} activation and $3\times3$ max-pooling.
Following the convolutional blocks, are two fully-connected layers with \emph{LeakyReLU} activation, and \emph{SoftMax} layer at the end.\footnote{LeakyReLU is defined as $f(x) = \max(0,x) + \alpha*\min(0,x)$, where $\alpha$ is a hyperparameter.}
We again use the MNIST dataset, and follow the same problem setup as in the first DNN problem by optimizing similar hyperparameters (full details in supplementary).

The third DNN problem incorporates a higher number of hyperparameters.
We increase the number of hyperparameters to $15$, $10$ continuous and $5$ integer, and use the same CNN and setup as in the second experiment. This test problem is referred as \textbf{15-CNN} in subsequent discussion.

Finally, we design the fourth DNN problem on the more challenging dataset, CIFAR-10, and in even higher dimensional hyperparameter space by optimizing $19$ hyperparameters, $14$ continuous and $5$ integer. This hyperparameter optimization problem is referred as \textbf{19-CNN} in subsequent discussions.
We optimize the hyperparameters of the same CNN network from Problem 15-CNN, except that we include four dropout layers, two in each convolutional block and two after each fully-connected layer.
We add the dropout rate of these four layers to the hyperparameters being optimized.

\paragraph{Baseline algorithms}
We compare HORD against Gaussian processes with expected improvement (GP-EI)~\cite{snoek2012practical}, Gaussian processes with predictive entropy search (GP-PES)~\cite{hernandez2014predictive}, the Tree Parzen Estimator (TPE)~\cite{bergstra2011algorithms}, and to the Sequential Model-based Algorithm Configuration (SMAC)~\cite{hutter2011sequential} algorithms.

\paragraph{Evaluation budget, trials and initial points}
DNN hyperparameter evaluation is typically computationally expensive, so it is desirable to find good hyperparameter values within a very limited evaluation budget.
Accordingly, we limit the number of optimization iterations to $200$ function evaluations, where one function evaluations involves one full training and evaluation of DNN.
We run at least five trials of each experiment using different random seeds.

The SMAC algorithm starts the optimization from a manually set Initial Starting Point (ISP) (also called default hyperparameter configuration).
The advantage of using manually set ISP as oppose to random samples becomes evident when optimizing high number of hyperparameters.
Unfortunately, both GP-based algorithms, as well as TPE, cannot directly employ a manually set ISP to guide the search, thus they are in slight disadvantage.

On the other hand, HORD fits the initial surrogate model on $n_{0}$ Latin hypercube samples, but can also include manually added points that guide the search in a better region of the hyperparameter space.
We denote this variant of HORD as HORD-ISP and test HORD-ISP on the 15-CNN and 19-CNN problems only.
We set the ISP following common guidelines for setting the hyperparameters of a CNN network (exact values in supplementary). We supply the same ISP to both HORD-ISP and SMAC for the 15-CNN and 19-CNN problems. HORD-ISP performed the same as HORD on the $6$ and $8$ hyperparameter cases. 

\paragraph{Implementation details }\label{sec:implement} We implement the HORD~algorithm with the open-source surrogate optimization toolbox pySOT~\cite{pysot}. HORD has a number of algorithm parameters including $n_0$ the number of points in the initial Latin hypercube samples, $m$ the number of candidate points, $w$ the cycling weights used in Eq.~\ref{eqn:weight} for computing candidate point score and the parameters controlling the changes in perturbation variances when there is no improvement. We have used the default values of these algorithm parameters as given in the pySOT software toolbox (since they were used in earlier RBF surrogate algorithms and tested on numerous benchmark optimization problems) and did not attempt to adjust them to improve performance.
We use the Spearmint library\footnote{\url{https://github.com/HIPS/Spearmint}} to obtain results for the GP-based algorithms.
We use the HyperOpt library\footnote{\url{https://github.com/hyperopt/hyperopt}}
to obtain results for the TPE method.
We use the public implementation of SMAC.
The HORD implementation, the networks being optimized, and other code necessary for reproducing the experiments are available at~\url{bit.ly/hord-aaai}.

\section{Numerical Results and Discussion}
\subsubsection{Algorithm Comparison Methodology}

A purpose to the algorithm comparison we have done is to assess which algorithms are more likely to be efficient and how does this efficiency vary with the number of hyperparameters and across different problems. We also would like to identify some of the possible causes for this variation. Table~\ref{tbl:results} shows the mean and standard deviation for the best test set error obtained by each of the methods after two hundred evaluations for all of the test problems. We see here that for $3$ out of $4$ problems HORD performs better than other algorithms as per the test set error. However, we analyze in detail, the comparative performance of algorithms in subsequent discussions via validation error, since it is the objective function in Eq.~\ref{eqn:hyp_opt_def}.  

\begin{figure}[t]
	\centering
	\includegraphics[width=1.0\linewidth, height=0.91\linewidth]{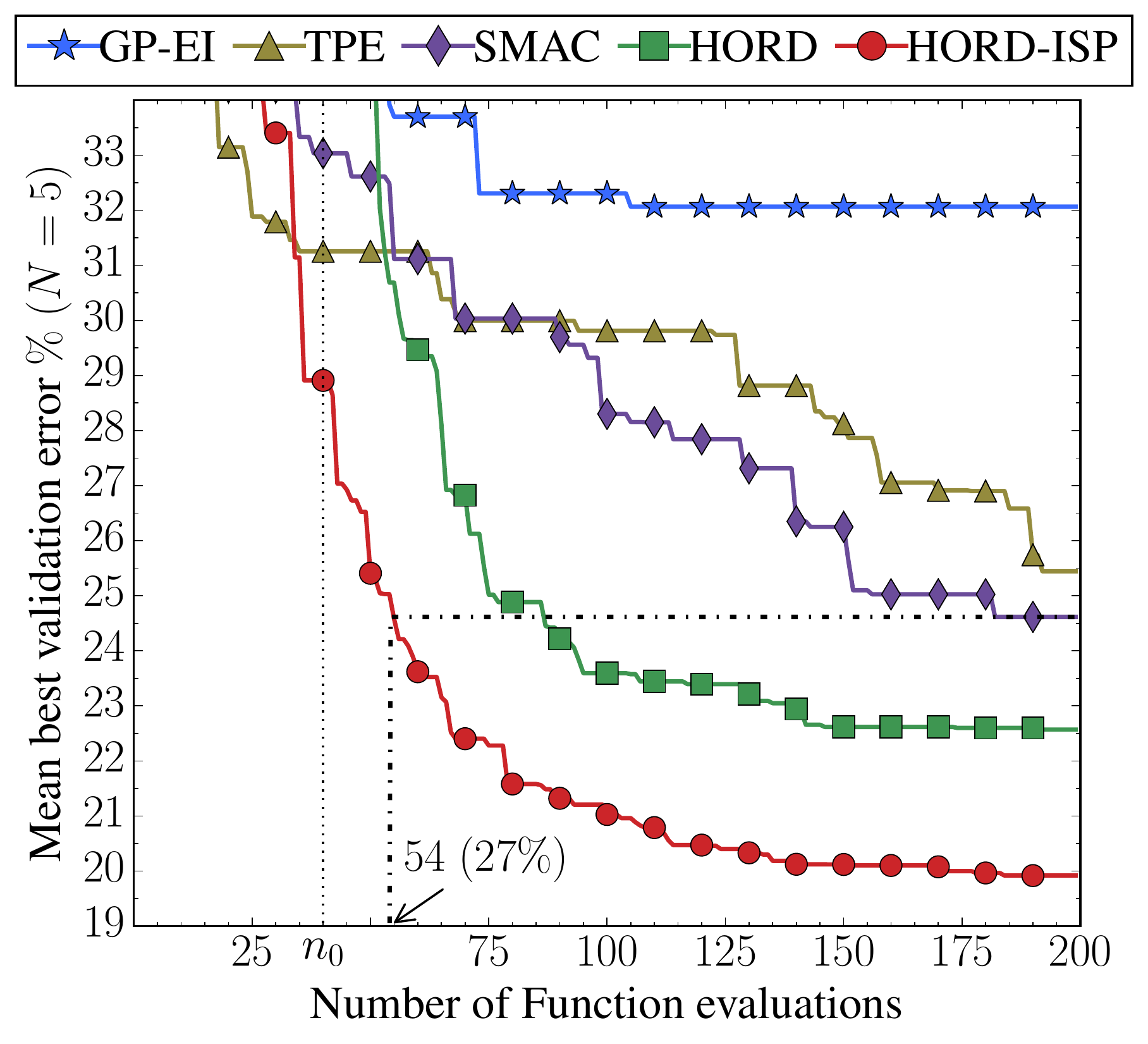}
   \caption{Efficiency comparison of HORD and HORD-ISP with baselines for optimizing a CNN with $19$ hyperparameters on the CIFAR-10 dataset (19-CNN). We plot  validation error curves of the compared methods  against  number of the function evaluations (averaged over $5$ trials). HORD and HORD-ISP show to be significantly more efficient than other methods. HORD-ISP only takes $54$ function evaluations to achieve the lowest validation error that the best baseline (SMAC) achieves after $200$ evaluations.  }
	\label{fig:eval_nets}
\end{figure}

Figure \ref{fig:eval_nets} plots the mean value of the best solution found so far  as a function of the number of expensive evaluations of $f(\mathbf{x})$ in Eq.~\ref{eqn:hyp_opt_def}.   The average is over five trials. 
Since we are minimizing, lowest curves are better.
We can see from the horizontal dotted line in Figure~\ref{fig:eval_nets},  that for the 19-CNN problem, HORD reaches in only $54$ evaluations the mean best validation error achieved by SMAC after $200$ evaluations, \textit{i.e}, HORD required only  $27\%$ of evaluations required by SMAC to get the same answer;  $27\%$ and $54$ are shown in lower right corner of Table~\ref{tbl:speedup}. Table~\ref{tbl:speedup} reports this percentage for HORD in comparison to all other algorithms and for all DNN hyperparameter optimization problems tested.  All the percentages are significantly less than $100\%$, implying that in every problem,  HORD obtained an equivalent mean answer to all other methods in less time.

Since HORD and all baseline algorithms  are stochastic, it is also important to do a more stringent test statistical test to  understand that if and after how many function evaluations (as a percentage of $200$) does HORD achieves best validation error (in multiple trials) that is statistically better than the best validation error (multiple trials) achieved by other algorithms after $200$ evaluations.
We employ the rank sum test for this analysis and report the percentages in Table~\ref{tbl:stats}. 
For 15-CNN and 19-CNN problems the comparative analysis reported in Table~\ref{tbl:speedup} and Table~\ref{tbl:stats} is against HORD-ISP.

To evaluate the performance of deep learning algorithms with optimized hyperparameters, we report the mean and standard deviation of best test set error for all algorithms and DNN problems after $200$ evaluations in Table~\ref{tbl:results}. Results indicate that HORD and HORD-ISP achieve the best results on average for all problems.

\begin{table}[t]
	\centering
	\caption{
		We report the mean and standard deviation (in parenthesis) of test set error obtained with the best found hyperparameter configuration after $200$ iterations in at least $5$ trials with different random seeds.
      The algorithm with the lowest test error is shown in bold.
	}
	{\fontsize{9.0pt}{10.0pt}\selectfont
		\setlength\tabcolsep{0.7mm}
		\begin{tabular}[b]{ccccc}
			\hline
			\noalign{\vskip 0.1em}
			Data Set & MNIST & MNIST & MNIST & CIFAR-10 \\
			\noalign{\vskip 0.1em}
			Problem &  $6$-MLP&   $8$-CNN  &  $15$-CNN & $19$-CNN  \\
			\hline
			\noalign{\vskip 0.1em}
			GP-EI       &$ 1.94 (.11) $&$ \mathbf{0.77(.07)} $&$ 0.99(.11) $&$  37.19(4.1)            $\\
			\noalign{\vskip 0.1em}
			GP-PES      &$    1.94(.07)          $&$    0.87(.04)          $&$      1.06(.07)        $&\textemdash\\
			\noalign{\vskip 0.1em}
			TPE         &$ 2.00 (.079) $&$ 0.96(.07) $&$ 0.97(.03) $&$ 27.13(3.2) $\\
			\noalign{\vskip 0.1em}
			SMAC        &$     2.13(.11)         $&$       0.85(.07)       $&$      1.10(.07)        $&$       29.74(2.1)       $\\
			\noalign{\vskip 0.1em}
			HORD        &$\mathbf{1.87 (.06)}$&$0.84(.04)$&$ 0.94(.07) $&$ 23.23(1.9) $\\
			\noalign{\vskip 0.1em}
			HORD-ISP &       \textemdash         &    \textemdash           &$\mathbf{0.82(.05)}$&$\mathbf{20.54(1.2)}$\\
			\hline
		\end{tabular}
		\label{tbl:results}
	}
\end{table}

\begin{table}[t]
	\centering
	\caption{
		The percentage of function evaluations (actual function evaluations are also reported in parenthesis) required by HORD (6-MLP, 8-CNN) and HORD-ISP (15-CNN, 19-CNN) to reach the mean best validation error achieved by other algorithms after 200 function evaluations.
	}
	{\fontsize{9.0pt}{10.0pt}\selectfont
		\setlength\tabcolsep{1.0mm}
		\begin{tabular}[b]{cllll}
			\hline
			\noalign{\vskip 0.1em}
			Data Set & MNIST & MNIST & MNIST & CIFAR-10 \\
			\noalign{\vskip 0.1em}
			Problem &  $6$-MLP&   $8$-CNN  &  $15$-CNN & $19$-CNN  \\
			\hline
			\noalign{\vskip 0.1em}
			GP-EI       &$ 78\%(155) $&$ 73\%(145) $&$ 18\%(36) $&$ 17\%(33) $\\
			\noalign{\vskip 0.1em}
			GP-PES      &$     16\%(32)        $&$      50\%(100)        $&$  10\%(20)            $&\textemdash\\
			\noalign{\vskip 0.1em}
			TPE         &$ 38\%(75)  $&$ 50\%(100)  $&$ 29\%(58)  $&$ 25\%(49)  $\\
			\noalign{\vskip 0.1em}
			SMAC        &$  20\%(39)             $&$  28\%(55)             $&$     20\%(40)          $&$   27\%(54)           $\\
			\noalign{\vskip 0.1em}
			\hline
		\end{tabular}
		\label{tbl:speedup}
	}
\end{table}

\begin{figure}
	\centering
	\includegraphics[width=1.0\linewidth, height=0.91\linewidth]{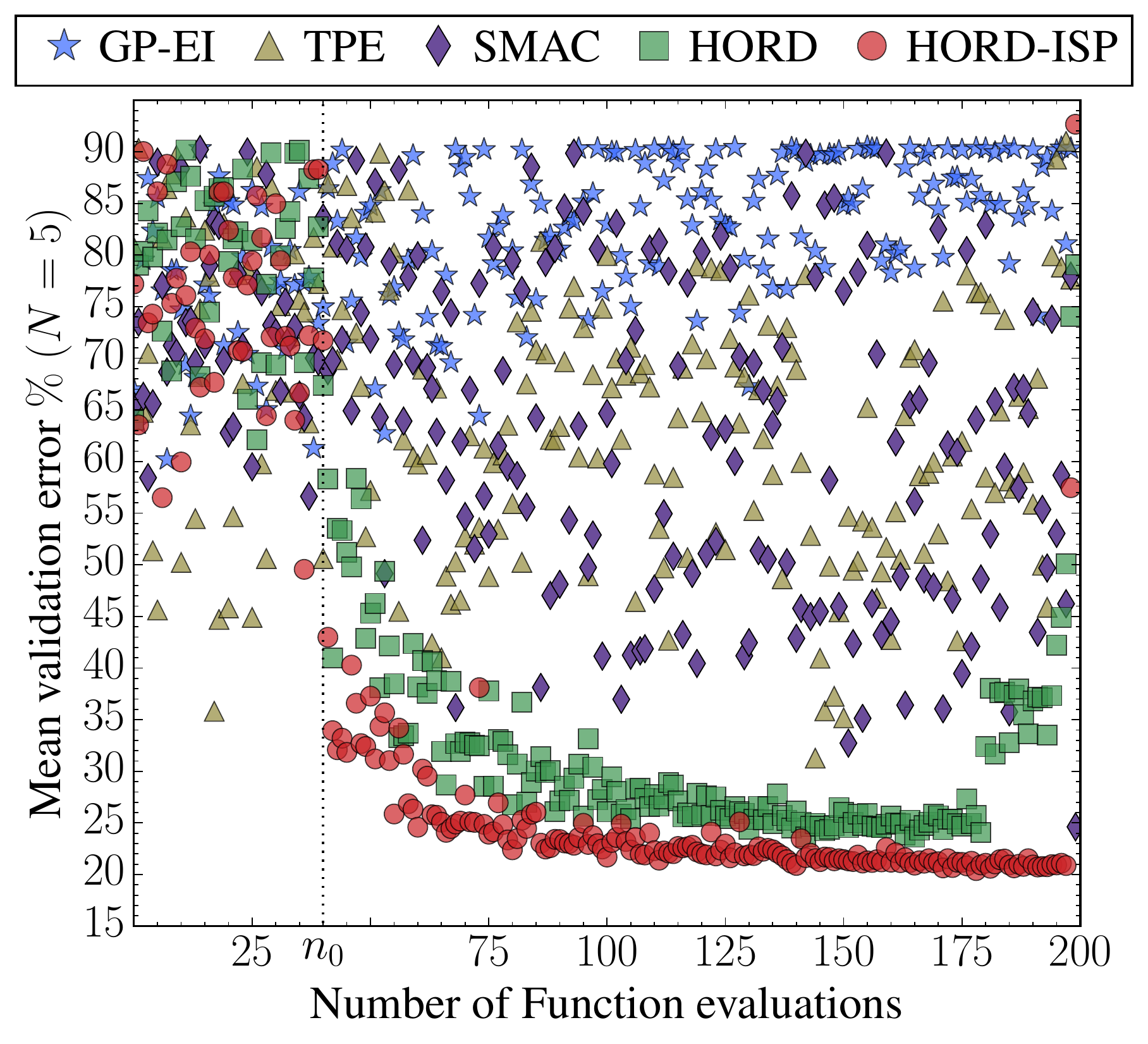}
	\caption{Mean validation error  \emph{v.s.}\ number of function evaluations  of different methods for optimizing 19 hyperparameters of CNN on CIFAR-10. One dot represents validation error of an algorithm at the corresponding evaluation instance. After $n_0$ evaluations, the searching of HORD and HORD-ISP starts to focus on the hyperparameters with smaller validation error ($\leq 35\%$), in stark contrast with other methods.}
	\label{fig:eval_nets2}
\end{figure}

\begin{table}[t]
	\centering
	\caption{
		We report the minimum percentage of function evaluations (actual function evaluations are also reported in parenthesis) required by HORD (6-MLP, 8-CNN) and HORD-ISP (15-CNN, 19-CNN) to obtain significantly (via Rank Sum Test) better best validation error than best error achieved by other algorithms after  $200$ function evaluations. If the difference in algorithms is insignificant at $200$ function evaluations, we report it as a ``--''. All statistical tests are performed at $5$ percent significance level. 
	}
	{\fontsize{9.0pt}{10.0pt}\selectfont
		\setlength\tabcolsep{1mm}
		\begin{tabular}[b]{ccccc}
			\hline
			\noalign{\vskip 0.1em}
			Data Set & MNIST & MNIST & MNIST & CIFAR-10 \\
			\noalign{\vskip 0.1em}
			Problem &  $6$-MLP&   $8$-CNN  &  $15$-CNN & $19$-CNN  \\
			\hline
			\noalign{\vskip 0.1em}
			GP-EI       &\textemdash &\textemdash &$ {29\%(58)} $&$ {18\%(35)} $\\
			\noalign{\vskip 0.1em}
			GP-PES      &$      {38\%(75)}        $&\textemdash &$  {14\%(27)}            $&\textemdash\\
			\noalign{\vskip 0.1em}
			TPE         &$ {67\%(134)}  $&$ {77\%(154)}  $&$ {62\%(123)}  $&$ {32\%(64)}  $\\
			\noalign{\vskip 0.1em}
			SMAC        &$  {39\%(77)}             $&$  {72\%(144)}             $&$     {29\%(58)}          $&$    {33\%(66)}           $\\
			\noalign{\vskip 0.1em}
			\hline
			
		\end{tabular}
		\label{tbl:stats}
	}
\end{table}

\subsubsection{Results for  Optimization of MLP and CNN network hyperparameters}

For the 6-MLP  problem we see that the average performance of HORD is better than all other algorithms after $200$ evaluations (see Column 2 of Table~\ref{tbl:speedup}). Furthermore, Table~\ref{tbl:stats} (Column 2) clearly indicates that HORD~found significantly better hyperparameters than the tree-based algorithms, SMAC and TPE, and GP-PES within $70\%$ of evaluations required by other algorithms after $200$ evaluations. The performances of HORD and GP-EI, however, are not statistically distinguishable (the $p$-value of rank sum test at $200$ function evaluations is $0.54$).
The performance of the GP-EI algorithm is expected since GP-based methods are well suited to low-dimensional optimization.
However, HORD~also performs almost exactly the same, while using about $1{,}000$ times less computing resources to propose new hyperparameter configurations. See supplementary for comparison of algorithm running times.

In the 8-CNN DNN case we again see that average performance of HORD is better than other algorithms after 200 function evaluations (see Column 3 of Table~\ref{tbl:speedup}). Furthermore, performance of GP-EI is statistically comparable to HORD ($p$-values of rank sum tests are $0.391$) after $200$ function evaluations. However, HORD statistically significantly outperforms the tree-based algorithms (see Column 3 of Table~\ref{tbl:stats}) and GP-PES, which shows that our algorithm can perform well on a more complex network and with slightly higher number of integer hyperparameters.

\subsubsection{Results for Optimization of CNN hyperparameters on MNIST and CIFAR-10}\label{sec:eval_sets}

The 15-CNN and 19-CNN experiments confirm that higher dimensional spaces are where HORD-ISP~truly shines (See Columns 4 and 5 of Table~\ref{tbl:speedup} and Table~\ref{tbl:stats}). As expected, the performance of the GP-based algorithms (see Column 5 of Table~\ref{tbl:stats}) degrades significantly in high dimensional search space.
While, HORD-ISP (with the same ISP as SMAC) continues to perform well and even outperforms the tree-based algorithms, SMAC and TPE, which are designed specifically for optimizing high number of hyperparameters.

An analysis in Figure~\ref{fig:eval_nets2}, indicates HORD and HORD-ISP are finding better solutions in most iterations than other algorithms for 19-CNN. We observe that after $n_{0}$ points are used to build the surrogate, HORD and HORD-ISP propose hyperparameter values that result in validation errors mostly in the range $20-30\%$. While, all the other algorithms are essentially performing a random search over the whole hyperparameter space. Furthermore, HORD and HORD-ISP after just few function evaluations from $n_{0}$, reach validation errors lower than the final lowest validation errors of all the other algorithms. We only compare HORD-ISP against GP, TPE and SMAC for 19-CNN. HORD-ISP obtains significantly betters results than other algorithms after $200$ evaluations, within $33\%$ of $200$ evaluations (see Column 5 of Table~\ref{tbl:stats}).

\subsection{Discussion}

The differences in results among the methods compared to HOD-ISP are dramatically different for the 15 and 19 dimensional hyperparameter problems. For example on 19-CNN problem, HORD-ISP  obtains in 33 function evaluations  the same average validation answer that GP-EI required 200 evaluations to obtain, meaning that HORD-ISP is about 6 (=200/33) times faster than GP-EI.  Even compared to SMAC, which is designed to work better in higher dimensions, HORD is about 3.7 (=200/54) times faster based on average times (Table 2).  When we use the more stringent statistical test (using 95\% significance level)  given in Table 3,  we see that for example HORD-ISP is over 3 (=200/66) times faster than the best of the other algorithms, which is SMAC (which also has an initial starting point, ISP). HORD without an initial guess was also much better than GP-EI or GP-PES in higher dimensions (see Figure~\ref{fig:eval_nets2}) when neither algorithm had an initial guess, both in terms of speed and in the quality of the answer HORD and HORD-ISP obtained after 200 evaluations.  For the lowest dimension problems as shown in Table 2, the speed up of HORD compared to other methods  is equal to or greater than 1.29 (=200/155) in comparison to the other methods. As table 3 shows some of the comparisons of HORD to the two GP methods on the lower dimensional problems are not statistically significant at the 95\% level because of variability among trials of both methods.  However, HORD (low dimension) and HORD-ISP (in higher dimensions) does get the best average answer over all dimensions and it is the only algorithm that performed well over all the dimensions.

To understand the differences in performance between the algorithms, we should consider the influence of  surrogate type and the procedure for selecting the next point $\mathbf{x}$ where the algorithm evaluates the expensive function $f(\mathbf{x})$. With regard to surrogate type,
comparisons have been made to the widely used EGO algorithm ~\cite{jones1998efficient} that has a Gaussian process (GP) surrogate and uses maximizing Expected Improvement (EI) to search on the surrogate.  EGO functions poorly at higher dimensions~\cite{regis2013combining}.
With a deterministic  RBF surrogate optimization method DYCORS was the best on all three multimodal problems in this same  paper. The previous comparisons are based on numbers of function evaluation. In the current study,  non-evaluation time was also seen to be much longer for the Gaussian Process based methods, presumably because they require a relatively large amount of computing to find the kriging surface and the covariance matrix, and that amount of computing  increases rapidly as the dimension increases.

However, there are also major differences among algorithms in the procedure for selecting the next expensive evaluation point based on the surrogate that are possibly as important as the type of surrogate used. All  of the algorithms compared  randomly generated candidate  points and select the one  point to expensively evaluate with a sorting  criterion  based on the surrogate and location of previously evaluated points. GP-EI, GP-PES, TPE and SMAC all distribute these candidate points uniformly over the domain.  By contrast HORD uses the strategy in LMSRBF ~\cite{Regis2007c} to create candidate points by generating them as perturbations (of length that is a normal random variable) around the current best solution. In ~\cite{Regis2007c} the uniform distribution  and the distribution of normally random perturbations around the current best solution were compared for  an RBF surrogate optimization and the LMSRBF  gave better results on a range of problems.  HORD uses the DYCORS strategy  to only perturb a faction of the dimensions, a strategy not used by the other 4 baseline methods.  This is a strategy that greatly improves efficiency of higher dimensional problems in other applications as shown in~\cite{regis2013combining}. So these earlier papers indicate that the combination of methods used in HORD to generate trial points and to search on them as a weighted average of surrogate value and distance has been a very successful method on other kinds of continuous global optimization problems. So it is not surprising these methods  also work well in HORD for the mixed integer problems of multimodal machine learning.

\section{Conclusion}\label{sec:conclusion}

We introduce a new  hyperparameter optimization algorithm HORD in this paper. Our results show that HORD (and its variant, HORD-ISP) can be significantly faster (e.g up to 6 times faster than the best of other methods on our numerical tests) on higher dimensional problems. HORD is more efficient than previous algorithms because it uses a deterministic Radial Basis surrogate and because it in each iteration generates candidate points in a different way that places more of them closer to the current best solution and reduces the expected number of dimensions that are perturbed as the number of iterations increase.   
 We also present and test HORD-ISP, a variant of HORD with initial guess of the solution. We demonstrated the method  is  very effective for optimizing hyperparameters of deep learning algorithms.

In future, we plan to extend HORD to incorporate parallelization and mixture surrogate models for improving algorithm efficiency. The potential for improving efficiency of RBF surrogate based algorithms via parallelization is depicted in \cite{Krityakierne2016} and via mixture surrogate models is depicted in \cite{Mueller2014b}.     

\section{Acknowledgements}
The work of Jiashi Feng was partially supported by National University of Singapore startup grant R-263-000-C08-133 and Ministry of Education of Singapore AcRF Tier One grant R-263-000-C21-112. C. Shoemaker and T. Akhtar were supported in part by NUS startup funds and by the E2S2 CREATE program sponsored by NRF of Singapore.

\bibliographystyle{aaai}
\bibliography{ilievski}

\end{document}